\documentclass[sigconf, nonacm]{aamas}

\usepackage{balance} 
\usepackage{amsmath} 
\usepackage{makecell}

\usepackage{xcolor}
\usepackage{todonotes}
\usepackage[normalem]{ulem} 

\usepackage{subcaption}

\usepackage{graphicx}
\usepackage[outdir=./images/]{epstopdf}

\usepackage{hyperref}
\usepackage{color}

\usepackage{amsmath}
\usepackage[ruled,linesnumbered]{algorithm2e}

\usepackage{amsfonts}

\usepackage{subcaption} 
\usepackage{graphicx} 

\usepackage{float} 

\usepackage{todonotes}

\newcommand{\AOA}[1]{\textcolor{black}{#1}}

\usepackage[most]{tcolorbox}
\newtcolorbox{generictextbox}[1]{
  colback=gray!5!white,
  colframe=gray!50!black,
  title={#1}, 
  boxrule=0.5pt,
  arc=3pt,
  left=6pt, right=6pt, top=4pt, bottom=4pt
}

\everypar{\looseness=-1}

\setcopyright{ifaamas}
\acmConference[AAMAS '26]{Proc.\@ of the 25th International Conference
on Autonomous Agents and Multiagent Systems (AAMAS 2026)}{May 25 -- 29, 2026}
{Paphos, Cyprus}{C.~Amato, L.~Dennis, V.~Mascardi, J.~Thangarajah (eds.)}
\copyrightyear{2026}
\acmYear{2026}
\acmDOI{}
\acmPrice{}
\acmISBN{}

\acmSubmissionID{634}

\title[AAMAS-2026 Formatting Instructions]{Ontological grounding for sound and natural robot explanations via large language models}

\author{Alberto Olivares-Alarcos}
\affiliation{
  \institution{Institut de Robòtica i Informàtica Industrial, CSIC-UPC}
  \city{Barcelona}
  \country{Spain}}
\email{aolivares@iri.upc.edu}

\author{Muhammad Ahsan}
\affiliation{
  \institution{Institute of Electrical and Control Engineering, National Yang Ming Chiao Tung University}
  \city{Hsinchu}
  \country{Taiwan}}
\email{muhammadahsanfs.ee12@nycu.edu.tw}

\author{Satrio Sanjaya}
\affiliation{
  \institution{Institute of Electrical and Control Engineering, National Yang Ming Chiao Tung University}
  \city{Hsinchu}
  \country{Taiwan}}
\email{satriosanjaya.ee11@nycu.edu.tw}

\author{Hsien-I Lin}
\affiliation{
  \institution{Institute of Electrical and Control Engineering, National Yang Ming Chiao Tung University}
  \city{Hsinchu}
  \country{Taiwan}}
\email{sofin@nycu.edu.tw}

\author{Guillem Aleny\`a}
\affiliation{
  \institution{Institut de Robòtica i Informàtica Industrial, CSIC-UPC}
  \city{Barcelona}
  \country{Spain}}
\email{aolivares@iri.upc.edu}

\begin{abstract}
Building effective human-robot interaction requires robots to derive conclusions from their experiences that are both logically sound and communicated in ways aligned with human expectations. This paper presents a hybrid framework that blends ontology-based reasoning with large language models (LLMs) to produce semantically grounded and natural robot explanations. Ontologies ensure logical consistency and domain grounding, while LLMs provide fluent, context-aware and adaptive language generation. 
The proposed method grounds data from human-robot experiences, enabling robots to reason about whether events are typical or atypical based on their properties. We integrate a state-of-the-art algorithm for retrieving and constructing static contrastive ontology-based narratives with an LLM agent that uses them to produce concise, clear, interactive explanations. 
The approach is validated through a laboratory study replicating an industrial collaborative task. Empirical results show significant improvements in the clarity and brevity of ontology-based narratives while preserving their semantic accuracy. 
Initial evaluations further demonstrate the system’s ability to adapt explanations to user feedback. 
Overall, this work highlights the potential of ontology–LLM integration to advance explainable agency, and promote more transparent human-robot collaboration.\looseness=-1
\end{abstract}


\newcommand{\BibTeX}{\rm B\kern-.05em{\sc i\kern-.025em b}\kern-.08em\TeX}

\begin{document}


\pagestyle{fancy}
\fancyhead{}


\maketitle 


\section{Introduction}
Bidirectional communication between interactive agents is a fundamental requirement for successful social coordination, particularly in collaborative human-robot scenarios~\cite{doi:10.1126/scirobotics.abm4183}. The emergence of large language models (LLMs) and other foundation models offers new opportunities to enable interactive information exchange between humans and robots~\cite{ZHANG2023100131}. For example, those models may contribute to the field of explainable agency (explaining the reasoning behind the behavior of goal-driven agents and robots)~\cite{10.5555/3306127.3331806}. However, foundation models often lack awareness of domain-specific knowledge and internal robot states, and their tendency to generate fluent yet ungrounded text can lead to hallucinations, misleading outputs, and false information. This is especially crucial in long-horizon tasks, where robots are expected to work for extended periods of time and the quantity of internal knowledge is huge. Retrieval-augmented generation (RAG) emerges as a promising solution to this drawback~\cite{gao2024retrievalaugmentedgenerationlargelanguage,10.1145/3637528.3671470}, using domain-specific knowledge sources to improve the reliability of explanations by aligning them with the contextual knowledge of the robot.\looseness=-1

Knowledge representation formalisms, such as ontologies, are great sources of domain-specific knowledge. Ontologies provide a formal and semantically rich representation of knowledge, making them a natural fit for the retrieval process within RAG-based explainable systems. Indeed, some authors argue that explainable systems must rest on explicit and reliable representations, which are only attainable through robust semantic and ontological frameworks. In essence, explanation presupposes semantics, and semantics presupposes ontology~\cite{GUIZZARDI2024102325}. Such an argument is empirically supported by a solid literature on ontology-based explainable artificial intelligence systems~\cite{10.1007/978-3-031-43789-2_33,10.1007/978-3-031-60606-9_15,Naqvi2024}. 

In the robotics domain, the development and use of ontologies have experienced stable growth for more than a decade~\cite{6385518,7886384,8593856,olivares-alarcos_alberto_Review2019,Borgo2019,bessler2020formal,Umbrico2020,9535344,app142411489}. Recent works have even investigated the generation of ontology-based robot explanatory narratives, claiming positive insights and also pointing out some limitations (e.g. ontology-based narratives tend to be verbose and complex to read)~\cite{10161359,aamas2024aoa}. Hence, the time is ripe to embrace ontology-informed language models as a foundation for explainable robots. The combination of LLMs, RAG and ontologies may create a robust framework for explainable robots, where LLMs generate content, RAG maintains reliability, and ontologies provide the semantic backbone, ensuring that explanations are grounded and semantically coherent.\looseness=-1


\begin{figure}[t!]
\centering
\includegraphics[width=1.00\columnwidth]{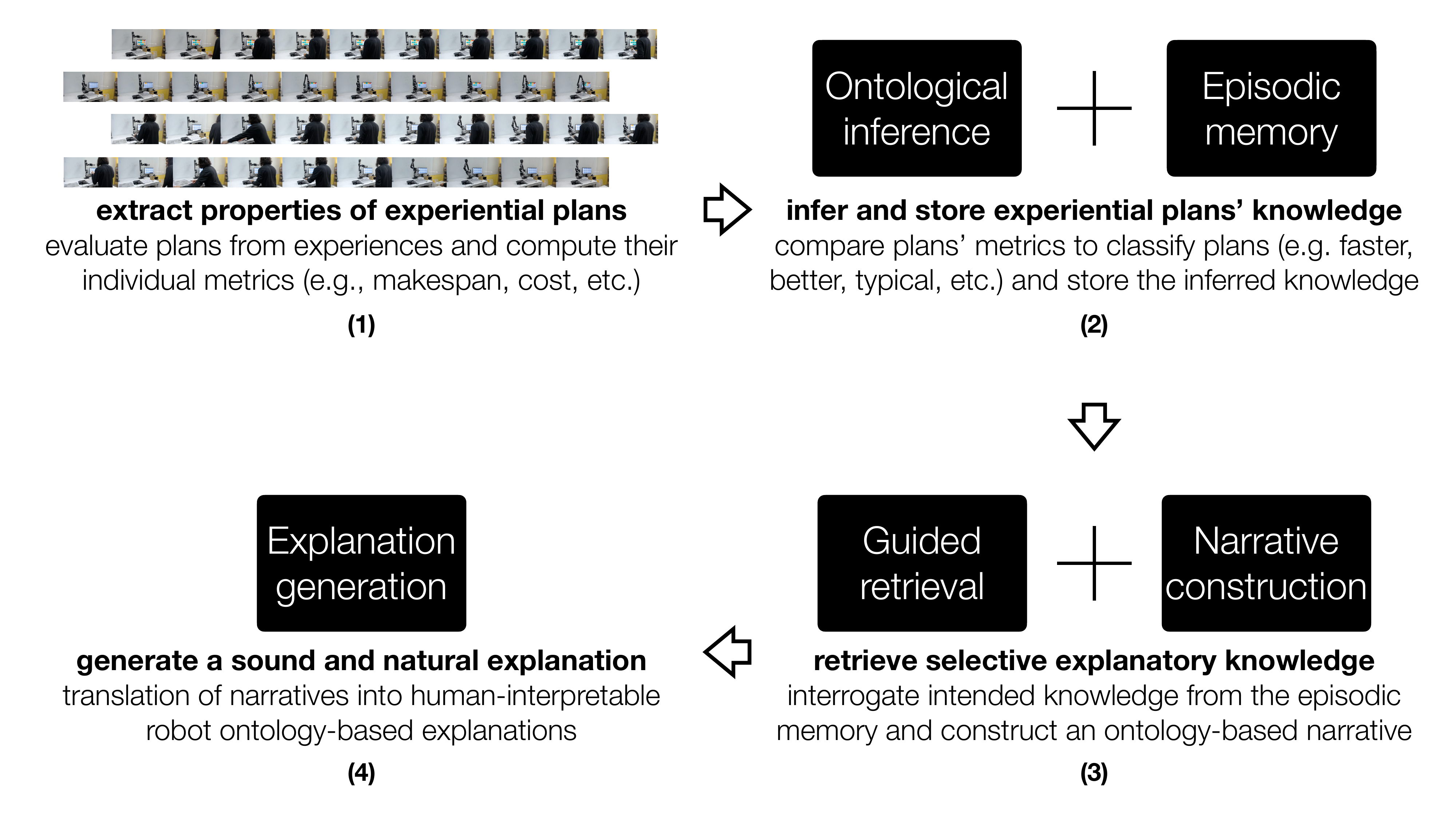}
\caption{Overview of the methodology for generating sound and natural explanations blending ontologies and LLMs. \AOA{For contrastive explanations (e.g., plan comparison),} (1) and (2) operate on $n$ plans, while (3) and (4) focus on $2$ plans (e.g., current and previous executions, current and prototypical, ...).\looseness=-1} 
\label{fg:methodology_overview}
\Description[<short description>]{<long description>}
\end{figure}

Building on these considerations, this work investigates the following research question: \textit{How can robots model and reason about recurring patterns in their experiences and generate explanations that compare them in a sound and natural manner?} This article addresses the question by presenting a novel methodology (see Fig.~\ref{fg:methodology_overview}) for generating robot ontology-based explanations. The methodology integrates a method to analyze data from robot experiences and extract common execution plan properties, a framework to generate robot memories of the extracted information, a state-of-the-art approach to retrieving \textit{sound} ontology-based robot experience narratives~\cite{olivaresalarcos2025ontologicalfoundationscontrastiveexplanatory}, and an LLM agent for generating \textit{natural} robot explanations of how the experiences compare. 
Our approach is validated with robot experiences collected using a laboratory mock-up task of a real industrial scenario in which a human and a robot collaborate to inspect the surface of solid-state drive (SSD) cases. The generated explanations, which compare the different experiences, are evaluated against the state-of-the-art method for ontology-based narratives (baseline). The performance of both methods is statistically compared using a set of objective evaluation metrics: explanation length (brevity), readability score (clarity), and semantic similarity (semantic coherence). The proposed method outperforms the baseline, producing shorter and clearer explanations while preserving the semantic meaning and coherence of the original narratives. It is also discussed how the novel methodology can adapt to user interactions asking for changes in the generated explanation: a shorter or clearer one. 
 
\section{Related work}
\label{sc:related_work}
\textbf{Sound formalisms for explainable robots:} 
Flores et al.~\cite{Flores2018}, proposed a method to transform the task history record of a robot, which is written in a declarative logical language for robot programming (SitLog), into a narrative knowledge representation. 
The literature also comprises some efforts to verbalize or narrate robot tasks knowledge encoded using symbolic planning formalisms (e.g. PDDL, RDDL)~\cite{rosenthal2016verbalization,Canal_AAAI2022}. The narratives described a plan's sequence, the rationale to include a task in the plan, etc. 
Sridharan’s work~\cite{sridharan2023}, combined non-monotonic logic, capable of supporting abductive reasoning, with deep learning, exploiting their complementary strengths in representation, reasoning, and learning to implement explainable agency. 
Some works leveraged the structure of ontology-based robot episodic memories (the collection of past personal experiences that occurred at particular times and places~\cite{tulving1972}) to retrieve past experiences knowledge to construct sound robot explanatory narratives~\cite{10161359,olivaresalarcos2025ontologicalfoundationscontrastiveexplanatory}. 
Some researchers exploited behavior trees, a mathematical model of plan execution, to produce robot explanations. Han et al.~\cite{10.1145/3457185}, framed the specific format of behavior trees into a new structure comprising the goal of the behavior tree, a set of subgoals, abstract steps to achieve the subgoals, and the final atomic actions. Depending on the target question to answer, a set of template-based explanations are generated. Love et al.~\cite{love2025temporalcounterfactualexplanationsbehaviour}, presented a method capable of generating causal and counterfactual explanations in response to contrastive ‘why’ questions to clarify the rationale behind robot decisions and actions. 
Those works rely on the natural robustness and interpretability of their respective formalism to produce sound narratives or explanations. However, while the generated narratives can be easily interpreted by experts (e.g. actual robot designers), it may not be so easy for final users. In most cases, the narratives contain truthful information, but in a verbose and complex format that hinders readability and understandability. Furthermore, once the narrative is generated, there is no possible interaction between the explainee (explanation's receiver) and the explainer (e.g. asking for more details).

\textbf{LLMs for explainable robots:} 
The literature is currently rich in works on generating robot explanations using LLMs. Often, those works use RAG architectures to manipulate information that is relevant for the execution of robot tasks to prompt LLM-based systems to produce domain-informed explanations. Those works were a significant source of inspiration for us. 
An et al.~\cite{10.5555/3709347.3743884}, proposed a logic-based explanation framework for Monte Carlo Tree Search, addressing its interpretability challenges. The framework translates user queries into logic and variable statements to ensure consistency with the underlying Markov Decision Process. 
LeMasurier et al.~\cite{10731331}, discussed a framework that to produce explanations using LLMs that are fed with the robot's behavior trees using the intermediate format proposed in~\cite{10.1145/3457185}. Retrieval approaches leveraging robot logs~\cite{sobrínhidalgo2024explainingautonomyenhancinghumanrobot} have been proposed to manage the limitations of long contexts, and have been further strengthened by incorporating a Visual Language Model (VLM)~\cite{sobrínhidalgo2024enhancingrobotexplanationcapabilities}. In other works, researchers use episodic information from robot past experiences in consonance with LLMs to generate explanations of behaviors, decisions and failures~\cite{liu2023reflectsummarizingrobotexperiences,anwar2025remembr,11203157,11203101}. \AOA{Although these works present interesting insights, the episodic information used in all of them lacks a generalizable and modular structure, an issue that ontologies would address.} Finally, some works combine robots' internal knowledge coming from different robot modules (task model, world model, causal model, episodic memory, etc.) with LLMs, to answer questions about robots' knowledge and generate explanations~\cite{bustamante2025raccoon,love2026hexarhierarchicalexplainabilityarchitecture}. Our work fits those larger frameworks, since our ontology-based robot knowledge may become one of their robot modules.\looseness=-1 



\section{Blueprint for robot ontology-based explanations}
\label{sc:method}
This work presents a novel methodology for robot ontology-based explanations consisting of four main steps (see Fig.~\ref{fg:methodology_overview}). First, we propose to evaluate human-robot collaborative experiences to extract the properties of the executed plans. Second, the extracted properties are used for ontology-based comparison and classification of plans, and the inferred knowledge is stored in an ontology-based episodic memory. Third, knowledge comparing plans in pairs is retrieved as ontological narratives. Fourth, the semantic content is refined to produce an explanation using an LLM agent. In this section, the aim is to provide some theoretical background that inspired our proposal and to discuss the details of the methodology design and implementation.  

\subsection{Background}
\textbf{Background on explainable agency:} 
The literature on explainable agency offers a well-established functional roadmap for designing explainable agents around four core functionalities~\cite{langley2024explainable}: (1) evaluating different candidate solutions to a task, (2) systematic indexing and storage of evaluation insights, (3) retrieval of stored knowledge, and (4) transmission of retrieved knowledge. 
First, during problem solving, the agent must evaluate multiple candidate solutions, analyzing their properties, so that it is possible to discriminate between them. 
Second, the agent must systematically index and store detailed records of its evaluation and classification in a structured repository, such as an episodic memory. Third, the agent must transform queries into retrieval cues to access relevant information from its memory. Fourth, once the information is retrieved, the agent translates it into human-interpretable language. 

Inspired by these ideas, we propose a methodology that tailors these four functionalities or steps to the semantic richness of human-robot interactions (see Fig~\ref{fg:methodology_overview}). Our proposal comprises: (1) robot experiences analysis and computation of their individual plan's properties, (2) ontology-based episodic storage and inference for plan comparison and classification, (3) selective and systematic retrieval of knowledge-based narratives, and (4) LLM-based generation of the final explanation for its transmission.\looseness=-1

\textbf{Insights from the social sciences:} 
There exists extensive research in philosophy, psychology, and cognitive science investigating how people define, generate, select, evaluate, and present explanations, showing that cognitive biases and social expectations shape the explanation process. Miller~\cite{MILLER20191} argues that the field of explainable artificial intelligence (XAI) can build on this existing research. Miller reviewed key works from philosophy, cognitive psychology/science, and social psychology, highlighting major findings and discussing how they can be integrated into XAI research and practice. The most important findings state that explanations are \textit{contrastive}, \textit{selected}, and \textit{social}. They are contrastive because they respond to counterfactual questions such as why one plan is better than another. Explanations are selective, typically presenting only part of the reasons, drawn from broader knowledge according to specific criteria. Finally, explanations are social, transferring knowledge in conversational form as part of interactions between agents. 
These qualities are inherent in the nature of the explanations themselves. Hence, being contrastive, selected and social are not just desirable but essential for the design and development of explainable robots. Note that there is a relationship between these properties and the functionalities (steps) discussed earlier. Contrastive explanations require the analysis of the qualities of multiple alternative options (functionality 1). The (biased) selection of explanation content depends on the knowledge storage and retrieval (functionalities 2 and 3), and can be learned from users' preferences. Finally, social explanations demand transmission in a human-interpretable form (functionality 4). 

Inspired by these ideas, this work proposes a methodology to produce \textit{contrastive} and \textit{social} (interactive) explanations that are \textit{selected} according to different levels of specificity or detail.

\subsection{Experiential plans' properties extraction}
In the first step of the proposed methodology, each plan derived from collaborative experiences between humans and robots is analyzed, focusing on finding the plans' properties. Note that these experiences contain variability and comprise different alternatives to perform the collaborative task, since humans do not always behave in a repetitive manner. Hence, the experiences are individually studied, extracting and computing the properties of the different executed plans. We specifically focused on three main aspects of plans: cost, makespan, and number of tasks. We are using the same plan qualities defined in a literature work that is later used as the evaluation baseline~\cite{olivaresalarcos2025ontologicalfoundationscontrastiveexplanatory}. We note that the approach allows us to consider other qualities, e.g. the workload distribution between human and robot. However, for our purpose, it is enough to stick to these. Note that the actual computation of some of the properties is application-dependent, for instance, the cost might be related to different aspects: energy consumption, the existence in the plan of complex or unsafe tasks, etc. In our case, cost was related to how much human intervention was needed (see Sec.~\ref{sc:evaluation_procedure} for more details).\looseness=-1

\subsection{Plans' knowledge inference and storage}
In the second step of our methodology, robots use the semantic structure of an ontology to infer knowledge about the plans' properties comparison and classification, and store it in a knowledge-based episodic memory (i.e., a time-indexed knowledge base). 

\subsubsection{Comparing plans' properties and plans}
The ontology for modeling the properties of plans defined in~\cite{olivaresalarcos2025ontologicalfoundationscontrastiveexplanatory}, is grounded using the information extracted in the previous step. Then, a set of logic rules, also defined in that work, is used to infer the comparison of the different experiences: e.g., plans with larger cost are defined as more expensive than other ones with lower cost.

\subsubsection{Classifying plans' properties and plans}
We intend to further reflect on the extracted data, analyzing the properties of the different plans, and finding patterns that are useful for the robot to classify experiences (e.g. a subset of plans is more recurring than others). Specifically, we propose to classify the different properties of plans as typical or atypical depending on whether they are within the robot experiences' mode or not. One might argue that this can be done by predefining a threshold of typical values for each property. However, that would require prior knowledge about the properties' common values, hindering the flexibility of the approach. Instead, we prefer to automatically extract this threshold from the actual data using a well-known state-of-the-art method for this: the Highest Density Interval (HDI). In this case, the assumption is the availability of a set of experiences. HDI is the narrowest interval within a probability distribution that captures a given probability mass (or sample proportion), representing the most recurring values of a parameter. Unlike equal-tailed intervals, the HDI is a Bayesian measure that always includes the mode and ensures that all values inside the interval have a higher probability density than those outside. This makes it a more direct and intuitive way to express uncertainty in the value of a parameter. 

\textbf{A novel implementation for computing HDI:} 
The HDI is widely used in Bayesian analysis and robust descriptive statistics because it always contains the region of highest density. However, its computation requires to know the probability distribution of the robot's dataset. For this, it is usual to compute the empirical HDI, a non-parametric approximation based on sorted data, requiring no distributional assumptions. 

Given a sample $\{x_1,\dots,x_n\}$, containing the values of a property for $n$ plans, and a coverage (mass) $\alpha \in (0,1)$, the empirical HDI approximates the theoretical HDI by finding the shortest interval containing at least $\alpha n$ points:
\[
k = \lceil \alpha \cdot n \rceil, \quad
\widehat{I}_\alpha = [ x_{(i^*)}, x_{(i^* + k - 1)} ],
\]
where
\[
i^* = \arg\min_{1 \le i \le n - k + 1} \big( x_{(i+k-1)} - x_{(i)} \big).
\]

Alg.~\ref{alg:hdi} shows the pseudo-algorithm that we propose to compute the empirical HDI using a sliding window over a set of $n$ values, where $n$ is the number of plans (experiences). Note that since HDI is univariate, this method must be applied $m$ times (the number of plan's properties) to find the interval for each property (e.g., the number of tasks, cost, etc.). Then, every specific plan's property is classified as typical or atypical depending on whether its value is within the interval or not. To analyze the computational complexity of the algorithm, we pay attention to its parts. The empirical HDI algorithm consists of sorting the sample of size $n$: $O(n \log n)$, scanning with a sliding window: $O(n)$. Hence, the overall time complexity is:
\[
O(n \log n), \quad \text{with space complexity } O(n).
\]


\textbf{A new ontological model for typical plan classification:} 
The content related to whether a plan's property is typical or atypical is newly introduced in our work; thus it is necessary a novel set of ontological entities and rules for knowledge representation and reasoning. The intended model shall cover a new ontological scope: \textit{how can a plan and its qualities be classified (e.g. typical)?}.\looseness=-1

The new model builds upon the ontology introduced in~\cite{olivaresalarcos2025ontologicalfoundationscontrastiveexplanatory}, extending and reusing its existing structure. Consequently, it inherits its upper ontology, the DOLCE+DnS Ultralite (DUL) foundational ontology~\cite{Borgo2021}. In DUL, the way to classify an \texttt{Entity} is through a \texttt{Concept}, thus we propose new sub-classes of \texttt{Concept}: \texttt{Typical Plan Quality Value}, \texttt{Atypical Plan Quality Value}, \texttt{Typical Plan}, and \texttt{Atypical Plan}. The first two are used to \texttt{classify} plan qualities (e.g. \texttt{Plan Cost}), while the two remaining \texttt{classify} \texttt{Plans}. The four new ontological classes are formalized using the description logic version of the Web Ontology Language 2, OWL 2 DL. The new model is instantiated by grounding the result of the HDI computation on each of the properties of interest: number of tasks, makespan, and cost. 

\begin{algorithm}[t!]
\caption{Empirical HDI via Sliding Window} \label{alg:hdi}
\KwIn{
 Sample $X = \{x_1, \dots, x_n\}$, coverage $\alpha \in (0,1)$}
 \KwOut{Empirical HDI interval $[ \text{lo}, \text{hi} ]$}
Sort $X$ into $X_{(1)} \leq X_{(2)} \leq \dots \leq X_{(n)}$ \\
 $k \gets \lceil \alpha \cdot n \rceil$\\
$\text{best\_width} \gets \infty$, $\text{best\_start} \gets 0$ \\
\ForEach {$i \in (1, n - k + 1)$}{
    $\text{width} \gets X_{(i+k-1)} - X_{(i)}$\\
        \If{$\text{width} < \text{best\_width}$} {
            $\text{best\_width} \gets \text{width}$\\
            $\text{best\_start} \gets i$\\
        }
    }
$\text{lo} \gets X_{(\text{best\_start})}$\\
$\text{hi} \gets X_{(\text{best\_start} + k - 1)}$\\

\end{algorithm}

\textbf{Logical rule to infer whether a plan is typical or atypical:} 
Assume a consistent instantiated ontology $\mathcal{O}_i$ as a set of triples $\langle subject, predicate, object \rangle$, which encodes knowledge about the qualities of different plans and the classification of the qualities as typical or atypical. The next step is to infer which plans are typical. Here, the criterion to decide if a plan is typical is satisfied when all its qualities are typical (i.e. they have a typical value). This is, if every quality of a plan ($D$) \texttt{is classify by} \texttt{Typical Plan Quality Value}: 
\begin{center}
    $\forall Q \ \ \exists D \ \ \langle Q, rdf.type, dul.Quality \rangle \land \langle D, rdf.type, dul.Plan \rangle \land \langle Q, dul.isQualityOf, D \rangle \land \langle Q, dul.isClassifyBy, TypicalPlanQualityValue \rangle$
\end{center}
The previous logical expression reads as follows: all  qualities ($Q$) related to a plan ($D$), \textit{`are classified by'} $TypicalPlanQualityValue$. The predicate \textit{`rdf:type'}, which relates an individual with its class, is a standard predicate from the Resource Description Framework (RDF)~\cite{McBride2004}. When the previous logical expression is true, we can say that $D$ \textit{`is classified by'} $TypicalPlan$, and the knowledge indicating it would be asserted:
\begin{center}
    $\mathcal{O}_i \leftarrow \mathcal{O}_i \cup \langle D, dul.isClassifiedBy, TypicalPlan \rangle. $
\end{center}

In contrast, when at least one of the properties of a plan is atypical, the plan $D$ will be classified as atypical: 
\begin{center}
    $\mathcal{O}_i \leftarrow \mathcal{O}_i \cup \langle D, dul.isClassifiedBy, AtypicalPlan \rangle. $
\end{center}

These logical expressions were combined into a single rule to classify a plan as typical or atypical. The robot knowledge base used in this work is written in Prolog~\cite{clocksin2012programming}, thus the rule was implemented as a Prolog predicate and is accessible online.\footnote{https://github.com/albertoOA/know\_demo} The time complexity of our implementation is $O(1)$ (constant complexity) in the best case, and $O(m)$ in the worst case, where $m$ is the number of qualities of a plan. The space complexity is also independent of the input size $O(1)$. In practice, this rule is run for multiple plans, thus being $n$ the number of plans, the overall time complexity is: $O(n)$ for the best case and $O(n \cdot m)$ the worst case. The overall space complexity remains $O(1)$, regardless of the number of plans $n$ or qualities $m$. Note that the worst case will occur only when all the qualities of all the plans are typical, which is unlikely in our scenario, since HDI ensures that only a subset of qualities are classified as typical. 

\subsection{Selective explanatory knowledge retrieval}
The third step of our methodology deals with selecting two plans and retrieving the explanation content according to a specific criterion. Our method focuses on contrastive knowledge (comparing the two plans), and allows specifying the level of detail. For this, our methodology integrates a literature method, ACXON~\cite{olivaresalarcos2025ontologicalfoundationscontrastiveexplanatory}, an algorithm designed to retrieve knowledge about divergences between ontological entities (e.g., plans) and to generate textual contrastive explanatory narratives. It operates on four main inputs: a time-indexed knowledge graph (triples annotated with a time interval in which they hold), the ontological class(es) of the instances to narrate, the temporal locality (interval of interest), and the desired level of specificity. 
Although it was conceived to work with contrastive narratives about pairs of robot plans, ACXON is general enough to be applied to other ontologies and entity classes. Indeed, because it leverages the structure of the knowledge graph and the vicinity between entities, it can automatically retrieve the information about typical properties and plans that we have introduced in our work. Depending on the chosen specificity level, ACXON produces three types of contrastive narratives, where specificity determines the granularity of detail included in the narrative i.e., the number of knowledge tuples. The retrieved knowledge tuples are aggregated into a final textual narrative, using some rules commonly used by humans when asked to combine chunks of knowledge. Note that narratives only contain sound ontological knowledge extracted from the knowledge base plus some basic connectors such as `and', or `while'. The textual format of the retrieved knowledge fits perfectly our proposed architecture in which the knowledge is fed to an LLM agent. For a set of $n$ robot plans (experiences), a total of $\frac{n(n-1)}{2}$ different pairwise narratives could be retrieved in this step. In the following, we present an example narrative of specificity 3 that compares a pair of plans (X and Y).\looseness=-1  

\begin{generictextbox}{Example of an ontology-based narrative}
    \textbf{\textit{`Plan X'}} is cheaper plan than and is shorter plan than and is faster plan than and is better plan than \textbf{\textit{`Plan Y'}}. \textbf{\textit{`Plan X'}} has makespan \textit{`Plan X makespan'}; while \textbf{\textit{`Plan Y'}} has makespan \textit{`Plan Y makespan'}. \textbf{\textit{`Plan X'}} has number of tasks \textit{`Plan X number of tasks'}; while \textbf{\textit{`Plan Y'}} has number of tasks \textit{`Plan Y number of tasks'}. \textbf{\textit{`Plan X'}} has cost \textit{`Plan X cost'}; while \textbf{\textit{`Plan Y'}} has cost \textit{`Plan Y cost'}. \textbf{\textit{`Plan X'}} is classified by \textit{`TypicalPlan'}; while \textbf{\textit{`Plan Y'}} is classified by \textit{`AtypicalPlan'}. \textit{`Plan X makespan'} has value \textit{`28.20'}; while \textit{`Plan Y makespan'} has value \textit{`40.35'}. \textit{`Plan X number of tasks'} has value \textit{`12'}; while \textit{`Plan Y number of tasks'} has value \textit{`18'}. \textit{`Plan X cost'} has value \textit{`0'}; while \textit{`Plan Y cost'} has value \textit{`3'}. 
\end{generictextbox}

\subsection{Robot explanation generation}
In the fourth and last step of the proposed methodology, an LLM agent uses the retrieved narrative and translates it into a more human-interpretable format to be transmitted. Recall that the intended explanation shall be sound and natural, which we translate into three main explanation qualities: brevity, clarity, and semantic coherence. There might be many prompting strategies to produce the intended explanation, but exploring them would be an engineering effort that is out of the scope of our scientific work. Hence, we decided to start the agent with a simple prompt that is as follows:\looseness=-1 

\begin{generictextbox}{System prompt instructions sent to the LLM}
    You are an agent that based on a given ontology-based narrative, shall provide a new narrative that: (a) is shorter than the original, (b) uses an easier language than the original, and (c) keeps the semantic meaning of the original.
\end{generictextbox}

\begin{figure*}[ht!]
\centering
\begin{subfigure}[b]{0.40\textwidth} 
\includegraphics[width=\textwidth]{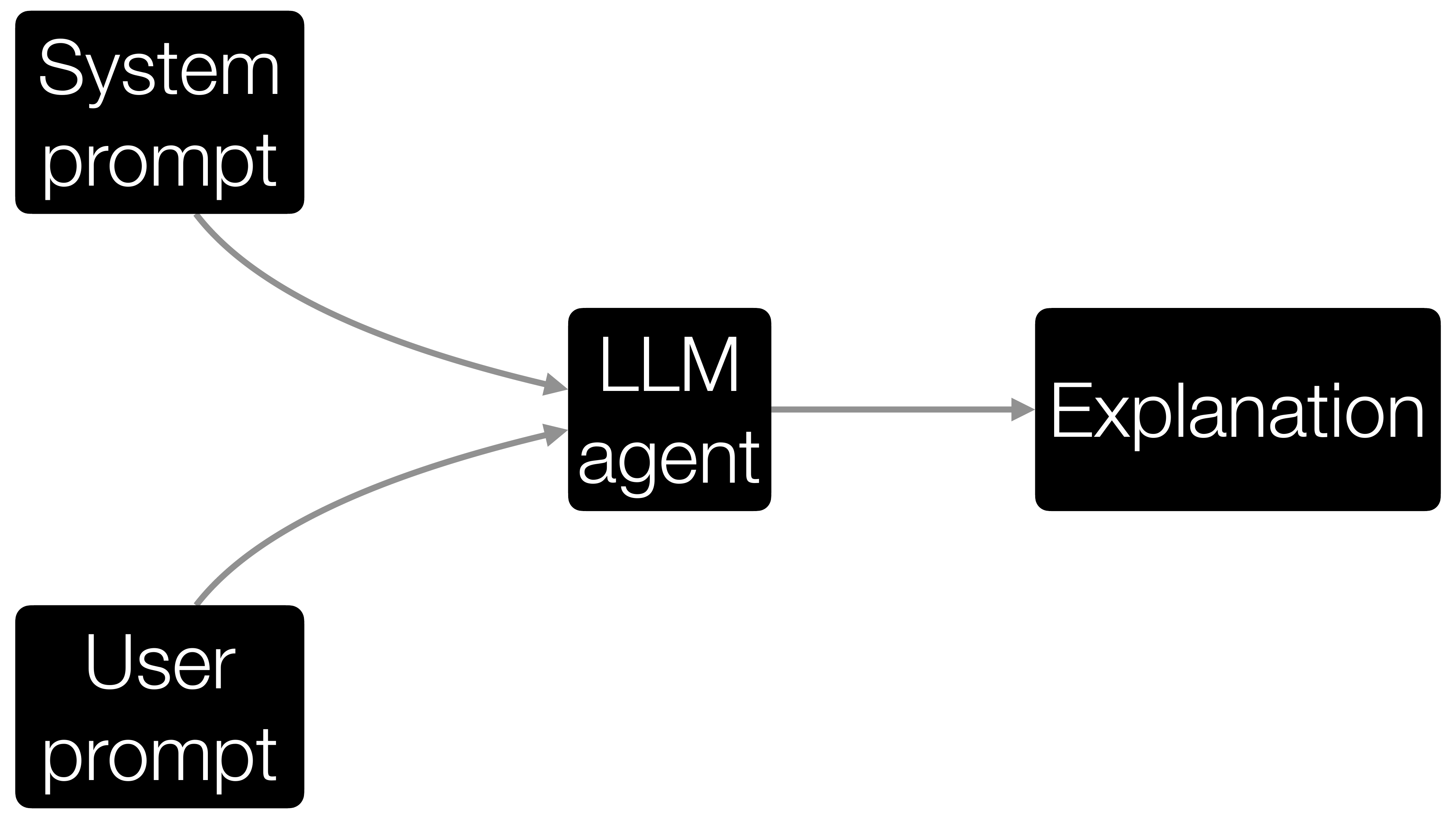}
\caption{}
\label{fg:simple_llm_configuration}
\end{subfigure} \hspace{0.1\textwidth}
\begin{subfigure}[b]{0.40\textwidth}
\includegraphics[width=\textwidth]{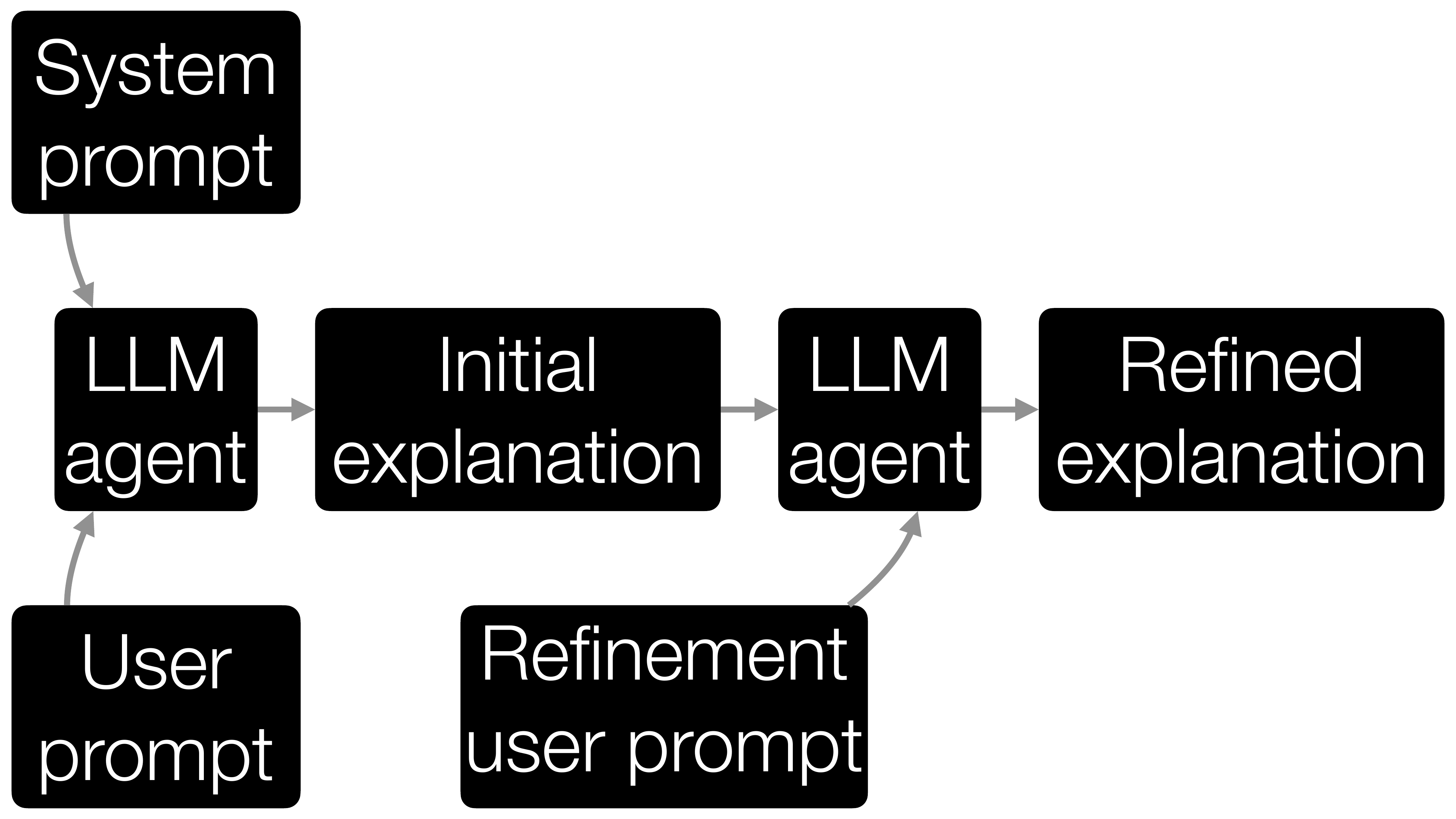}
\caption{}
\label{fg:interactive_llm_configuration}
\end{subfigure} 
\caption{LLM-based explanation generation settings. (a) A single system prompt and user prompt (ontology-based narrative) are used to generate the explanation. (b) Once the initial explanation is generated, a second user prompt is used to refine the explanation (e.g., asking for a shorter version).}
\label{fg:llm_configurations}
\Description[<short description>]{<long description>}
\end{figure*}

This prompt is sent to the agent as a system prompt, a set of initial instructions that provide the language model with context, defines its role, and establishes its behavioral guidelines for an interaction. Then, an ontology-based narrative comparing a pair of experiences (plans) is fed as a user prompt, so that the LLM agent returns the explanation. This process is repeated for each narrative comparing a pair of plans retrieved in the previous methodology's step, for $n$ plans, that would be one of the $\frac{n(n-1)}{2}$ narratives. 
The implementation\footnote{https://github.com/albertoOA/know\_rox} provides an environment that is later used to evaluate the approach (see Fig.~\ref{fg:simple_llm_configuration}), a setting where the agent provides an explanation given a system and a user prompt. In the following, we present the refined explanation using the ongoing example that compares the pair of plans (X and Y).\looseness=-1 

\begin{generictextbox}{Example of an LLM-based explanation}
    \AOA{\textbf{\textit{Plan X}} is cheaper, faster, shorter, and better than \textbf{\textit{Plan Y}}. \textbf{\textit{Plan X}} takes 28.20 time units, has 12 tasks, and costs 0. \textbf{\textit{Plan Y}} takes 40.35 time units, has 18 tasks, and costs 3. \textbf{\textit{Plan X}} is a typical plan; \textbf{\textit{Plan Y}} is an atypical plan.}
\end{generictextbox}

The LLM-based agent was deployed using \textit{Ollama},\footnote{https://ollama.com/} an open-source platform optimized for running large language models locally. Additionally, we employed \textit{Pydantic AI},\footnote{https://ai.pydantic.dev/} a Python-based agent framework designed for developing production-grade applications and workflows powered by generative AI.

\subsection{Robot explanation generation as interaction}
Blending the use of ontologies with the use of LLMs, our methodology is expected to effectively produce sound and natural explanations, improving the brevity and clarity of ontology-based narratives. Furthermore, the use of ontology-informed LLMs also brings a powerful interactive capability\AOA{, a requirement repeatedly demonstrated in social-science studies~\cite{MILLER20191}. For some structured scenarios, patterns of explanations or pre-formatted messages may be useful. However, in non-structured cases such as human-robot interactive scenarios, robot explanations must adapt to user goals, clarification requests, and evolving dialogue context.} In our case, after receiving an initial explanation, users might require a shorter version. They might also ask for specific details that were skipped when shortening the ontology-based narrative: e.g., the actual value of the cost of the two plans, not just if one is more expensive. For these cases, our methodology can be easily extended by slightly modifying the fourth and last step. Fig.~\ref{fg:interactive_llm_configuration} depicts an overview of the proposed modification. \AOA{Re-using the ongoing example, let's consider that the user asks: \textit{`Make the explanation shorter'}, our system will return a refined explanation.}\looseness=-1 

\begin{generictextbox}{LLM-based refined explanation after interaction}
    \AOA{\textbf{\textit{Plan X}} (typical) is better, cheaper, faster, and shorter than \textbf{\textit{Plan Y}}. It runs in 28.20 units, has 12 tasks, and costs 0, versus \textbf{\textit{Plan Y}}’s 40.35 units, 18 tasks, and cost 3.}
\end{generictextbox}



\section{Evaluation of robot ontology-based explanations}
\label{sc:evaluation}

\subsection{Collaborative SSD case inspection}
\begin{figure}[t!]
\centering
\includegraphics[width=1.00\columnwidth]{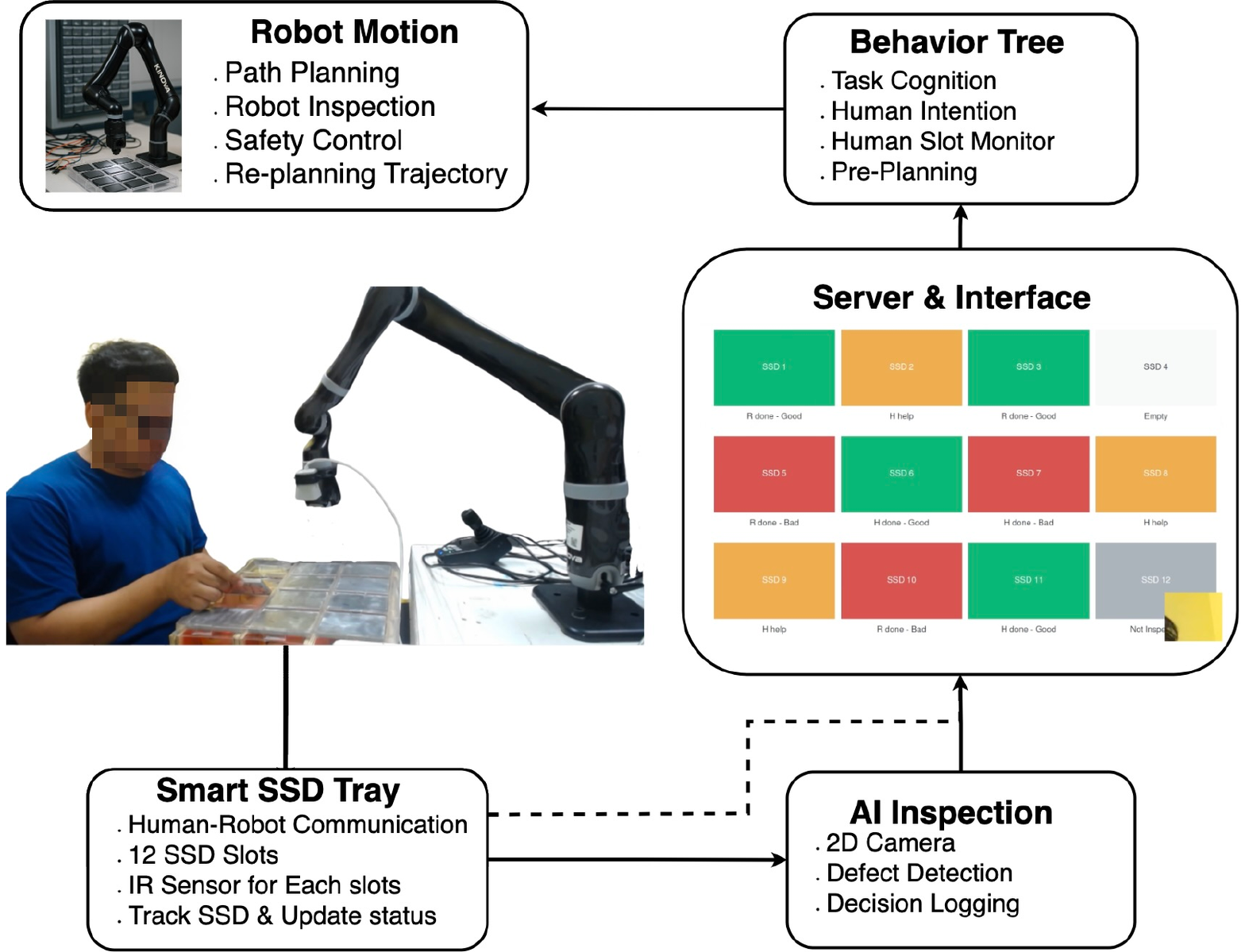}
\caption{Overview of the implementation for the collaborative SSD inspection task.} 
\label{fg:use_case_implementation}
\Description[<short description>]{<long description>}
\end{figure}

The evaluation of the proposed methodology is contextualized in a lab mock-up of a real industrial task, a human-robot collaborative inspection of the surface of metallic SSD cases. In this inspection task, there is a tray with 12 SSD units, and the robot helps the operator inspecting some of the units, although the robot might also try to perform the entire task. The robot is able to recognize when the precision of its inspection is low for a particular piece and ask the human for reassurance. The setup combines a 6-DOF robotic arm with a 2D camera for visual data capture, an AI-based defect detection model for real-time classification, and a user interface for efficient human-robot interaction. The implementation of the task comprises four core modules: a robotic manipulator for motion execution and trajectory planning, an AI-driven visual inspection module, a sensor-equipped smart SSD tray, and a behavior tree-based task execution model. Together, these modules enable synchronous collaboration between the human and the robot. A server-side graphical user interface (GUI) offers real-time visualization of task progress and inspection results. Note that communication from the robot to the human to indicate which cases need a second inspection occurs through the GUI. The complete implementation pipeline is shown in Fig.~\ref{fg:use_case_implementation}.

\subsection{Evaluation procedure and setup}
\label{sc:evaluation_procedure}

\subsubsection{Generating human-robot collaborative experiences}
In total, we collected 18 executions of the task, completing the inspection of 12 SSD cases, with a single human. To generate variability, we varied the confidence threshold of the robot's visual inspection and also the sequence of actions that the human followed. When the robot was uncertain whether one SSD case was damaged or not, i.e. classification confidence below the threshold, the case was left for the human to review. Data for each experience were collected using \textit{rosbags}, a file format and tool within the Robot Operating System (ROS) used to store ROS message data. The data includes the tray state during the whole task (to be inspected, damaged, correct), the sequence of actions of the human and the robot, and images from the robot's visual inspection.

\subsubsection{Experiential plans' properties extraction}
We manipulated the data from the collaborative SSD case inspection scenario, and extracted the knowledge that we considered relevant for describing the experiences: the number of tasks, the makespan, and the cost. The number of tasks includes all the SSD cases inspected by the human and the robot, which may be twelve or more, since the cases in which the robot is not confident are inspected twice. The makespan is calculated as the total time that it takes to complete the inspection of the entire tray. Finally, the cost is equal to the number of times that the robot doubted. HDI was computed for each property using a mass value of $0.68$, which means that at least 68\% of the plans' properties are classified as typical. This coverage or mass is often called \textit{one-sigma}, since given a data set approximately distributed normally, approximately 68\% of the data is within one standard deviation of the mean (according to the \textit{empirical rule}).\looseness=-1

\subsubsection{Plans' knowledge inference and storage}
We initialized a knowledge base, which works as an episodic memory, and asserted the abstraction of the data obtained in the previous step in the form of ontological triples. First, the knowledge about the comparison between plans' properties and plans. This includes knowledge such as:\looseness=-1
\[
\langle PlanX \ \ Cost, hasBetterQualityValueThan, PlanY \ \ Cost \rangle.
\]
We also asserted knowledge of whether the plan's qualities were typical or not: 
\[
\langle PlanX \ \ Cost, dul.isClassifyBy, TypicalPlanQualityValue \rangle.
\]

Once the knowledge was stored in the ontology-based episodic memory, we run the logical rules to infer how the different plans relate to each other, and also to classify plans as typical or atypical. The rules produced new knowledge, which was also asserted to the knowledge base: $\langle PlanX, ocra.isMoreExpensivePlanThan, PlanY \rangle$, and $\langle PlanX, dul.isClassifiedBy, TypicalPlan \rangle$.


\subsubsection{Selective explanatory knowledge retrieval}
Once there was an active knowledge base containing all the knowledge about how the different experiences or plans compare and are classified, we retrieved the content to be used in the explanations. This was generated using the state-of-the-art algorithm for ontology-based narrative generation~\cite{olivaresalarcos2025ontologicalfoundationscontrastiveexplanatory}. This algorithm takes four inputs: the time-indexed knowledge graph (the active knowledge), the ontological class(es) of the instances to compare is (\texttt{dul.Plan}), the temporal locality (a time interval (\_, Inf)), and the desired level of specificity (the three levels are used). Hence, we retrieve $\frac{n(n-1)}{2}$ narratives comparing in pairs the 18 plans ($n=18$) between each other, in total 153 narratives.\looseness=-1

\subsubsection{Robot explanation generation}
The retrieved narratives are sent to the LLM agent, which generates individual explanations using each of them. We used the \textit{gpt-oss:20b} model, a 20 billion-parameter version of openAI’s open-weight models designed for powerful reasoning, agentic tasks, and versatile developer use cases. We decided to use this model because it is openly available and because it provides a great trade-off between model size and accuracy. It was important for us that the model could be easily run in a single GPU since in the future we would like to be able to run our approach in real robot execution time. Note that other models can be easily integrated into our methodology.

\subsubsection{Computer setup for running the experiments}
The software for the test was run on a desktop PC with an Intel Core i9-11900KF 11th Gen CPU (8x 3.50 GHz), 64 GB DDR4 RAM, and an NVIDIA GeForce RTX 3090 GPU.

\subsection{Evaluation metrics}
\label{sc:metrics}
The evaluation was conducted using a set of objective metrics, selected to align with established practices in the state-of-the-art. The metrics aim to evaluate the three main intended features of our generated explanations: brevity, clarity, and semantic coherence.\looseness=-1

\textbf{Explanation length.} This metric was used to evaluate the brevity of the explanations. In this work, the metric is calculated as the number of words $\mathcal{N}_{w}$. Explanations are social, thus a good quality index is to measure how long they are, which is proportional to how much time it would take to communicate them.\looseness=-1 

\textbf{Readability score.} To evaluate the clarity of the explanations, we used the Flesch Reading Ease (FRES) readability $\mathcal{R}_{fres}$ metric~\cite{kincaid1975derivation}, proposed in the 1970s and still one of the best tools to evaluate how easy something is to read. The FRES test measures readability by analyzing word, syllable, and sentence counts. It computes the average number of words per sentence and syllables per word, assuming that shorter words and sentences enhance readability. Higher scores indicate easier-to-understand text. The maximum possible FRES score is 121.22, achieved when every sentence consists of a single one-syllable word. There is no defined lower bound, thus highly complex sentences may produce negative scores. We used the implementation of FRES from the Textstat Python library.\footnote{https://pypi.org/project/textstat/} Other readability metrics were considered and tried: automated readability index, Gunning fog index, Dale-Chall score~\cite{chall1995readability}, etc. However, results were similar, so we decided to use FRES because it is one of the most well-known metrics in the state-of-the-art.\looseness=-1

\textbf{Semantic similarity.} This metric is used to evaluate whether the process of transforming the initial ontology-based narrative into a shorter and clearer explanation results in a loss of semantic soundness. We used the cosine similarity $\mathcal{S}_{cs}$ function from the \textit{Scikit-learn} Python library,\footnote{https://scikit-learn.org/} which directly calculates the cosine similarity between two vectors or matrices, returning a value $\in (0,1)$. The retrieved ontology-based narrative and the explanation generated with the LLM agent were converted into a matrix of token counts using \textit{CountVectorizer} from the same library. In this process, we avoided counting most common words in English text, such as `and', `the', etc., which are presumed to be uninformative in representing the content of the text and which were removed to avoid them being used in measuring similarity between the two texts.\looseness=-1


\subsection{Results of evaluating the brevity, clarity and semantic coherence of the explanations}
\label{sc:results}

A statistical analysis was conducted to evaluate the significance of the improvements achieved by the proposed approach in terms of brevity and clarity with respect to the baseline, while preserving the semantic coherence of the ontology-based narratives. The two different methods were compared across three levels of specificity (independent variable) and assessed using the metrics introduced in Section~\ref{sc:metrics} (dependent variables). The normality of the difference between the results obtained by both methods was examined using the Shapiro-Wilk test ($\alpha=0.05$). For all cases, the normality assumption can be rejected ($p<0.05$), but since the distribution of the difference between the results was approximately symmetric and the number of samples was sufficiently large (N=153), parametric tests can be used. \textit{Paired-t test} was used for the \textbf{explanation length} and the \textbf{readability score}, while \textit{one-sample-t test} was applied for the \textbf{semantic similarity} metric. 

For the readability score at specificity 1, no significant difference was observed between the baseline and the proposed approach, although both yielded high readability scores. This outcome was expected, as baseline ontology-based narratives at this level are typically short and inherently easy to read. For the rest of the metrics and levels of specificity, the statistical results indicate that there is a significant large difference between the baseline and our method $p<0.001$. The average results of the evaluation metrics for the 153 pairs of plans, across both approaches and specificity levels, are summarized in Tab.~\ref{tab:average_results}. The detailed statistical analysis is reported Tab.~\ref{tab:statistical_results}. 

\begin{table}[t]
	\caption{Average evaluation results for each metric and the 153 pairs of plans. The mean difference is statistically significant \textit{p$<$0.001} for all the cases, with the only exception of $\mathcal{R}_{fres}$ for specificity 1.}
	\label{tab:average_results}
    \centering
	\begin{tabular}{c|ccc|ccc}\toprule
		\textit{\textbf{Method}}&\multicolumn{3}{c|}{\textbf{Baseline} (ACXON)}& \multicolumn{3}{c}{\textbf{Ours}} \\ 
		\textit{Specificity} &1&2&3 & 1&2&3 \\ \midrule 
 \textit{$\mathcal{N}_{w}$} &28& 741.7& 1788.6& \textbf{18.7}& \textbf{104.7}&\textbf{131.8} \\ 
 \textit{$\mathcal{R}_{fres}$}& \textbf{87}& 67.6& 48.2& 86.3& \textbf{80.7}& \textbf{84.6} \\ 
  \textit{$\mathcal{S}_{cs}$}&- &- &- & \textbf{0.7107}& \textbf{0.8668}& \textbf{0.7935} \\ 
	\end{tabular}
\end{table}

\begin{table}[t]
	\caption{Statistical evaluation results for each metric and specificity level. The specific statistical test metrics for each case is denoted between square brackets.}
	\label{tab:statistical_results}
    \centering
	\begin{tabular}{c|ccc}\toprule
		\textit{Specificity} &1&2&3 \\ \midrule
 \textit{$\mathcal{N}_{w}$ $[t(152),p]$} & $[31.1,< .001]$ & $[16.7,< .001]$ & $[14.7,< .001]$ \\
 \textit{$\mathcal{R}_{fres}$ $[t(152),p]$}& $[1.9, .062]$ & $[8.4, < .001]$ & $[15.4, < .001]$ \\ 
  \textit{$\mathcal{S}_{cs}$ $[t(152),p]$}& $[9.9,< .001]$ & $[31.8,< .001]$ & $[15.3,< .001]$ \\ 
	\end{tabular}
\end{table}

The performance of the proposed approach is consistently superior to that of the baseline ontology-based narratives, especially for the second and third levels of specificity. The proposed explanations are substantially shorter and clearer, while maintaining high semantic fidelity. Specifically, considering \textbf{explanation length}, our explanations are a 33\%, 86\% and 93\% shorter on average than the baseline explanations for the specificity level 1, 2, and 3, respectively. In the case of the \textbf{readability score}, the proposed method produces explanations that are a 19\% and 76\% easier to read and comprehend than the baseline narratives at levels 2 and 3. Importantly, this considerable gain in brevity and clarity does not come at the cost of \textbf{semantic coherence}. Indeed, the semantic cosine similarity was consistently above 0.7, indicating a strong semantic correlation between the generated explanations and the baseline. 

These results demonstrate that integrating ontological reasoning with LLM-based refinement enables the generation of concise yet semantically faithful explanations. The findings highlight the complementary roles of symbolic and neural representations: while the ontology guarantees logical consistency and grounding, the LLM enhances linguistic fluidity and contextual adaptation.

\subsection{Results of evaluating the interactive aspect of the explanations}
\label{sc:results_interactive}
\AOA{To validate the system's ability to adapt through dialogue, we conducted a statistical analysis of the explanation updates triggered by user interaction. While a comprehensive evaluation of all interactive capabilities of our method is inherently complex, we isolated a measurable interaction type to ensure empirical rigor: \textit{user requests for further brevity}. The \textbf{hypothesis} is that our method will be able to adapt, addressing the request and generating a shorter explanation, while maintaining the clarity and semantic coherence.} 

\AOA{The non-interactive and interactive versions of our method were compared across three levels of specificity (independent variable) and assessed using the metrics introduced in Section 4.3 (dependent variables). The normality of the difference between the results obtained by both methods was examined using the Shapiro-Wilk test ($\alpha=0.05$). As it happened before, the normality assumption can be rejected for all cases ($p<0.05$), but since the distribution of the difference between the results was approximately symmetric and the number of samples was sufficiently large (N=153), parametric tests can be used. \textit{Paired-t test} was used for the \textbf{explanation length}, the \textbf{readability score}, and for the \textbf{semantic similarity} metric.\looseness=-1} 

\AOA{The results demonstrate that the refined explanations were significantly shorter than the initial LLM-generated outputs ($p<0.001$), confirming successful adaptation. Note that this reduction in length did not compromise quality; readability results remained stable (non-significant differences), and the semantic similarity to the source ontology-based narrative showed only a small significant decrease. This validates our hypothesis, proving that the model effectively prunes information maintaining core meaning and without losing much linguistic clarity. The average results of the evaluation metrics for the 153 pairs of plans, across both approaches and specificity levels, are summarized in Tab.~\ref{tab:average_results_interactive}. The detailed statistical analysis is reported Tab.~\ref{tab:statistical_results_interactive}.}



\begin{table}[t]
	\caption{Average evaluation results for each metric and the 153 pairs of plans, for the comparison between our method with and without user interaction. The mean difference is statistically significant \textit{p$<$0.001} for all the cases of \textit{$\mathcal{N}_{w}$} and \textit{$\mathcal{S}_{cs}$}, and it is non-significant in all the cases for \textit{$\mathcal{R}_{fres}$}.}
	\label{tab:average_results_interactive}
    \centering
	\begin{tabular}{c|ccc|ccc}\toprule
		\textit{\textbf{Method}}&\multicolumn{3}{c|}{\textbf{Ours non-interactive}}& \multicolumn{3}{c}{\textbf{Ours interactive}} \\ 
		\textit{Specificity} &1&2&3 & 1&2&3 \\ \midrule 
 \textit{$\mathcal{N}_{w}$} &18.6& 110.4& 120.7& \textbf{17.2}& \textbf{63.4}&\textbf{61} \\ 
 \textit{$\mathcal{R}_{fres}$} & 86.3& 80.9& \textbf{87.2}& \textbf{87.7}& \textbf{81.2}& 85.6 \\ 
  \textit{$\mathcal{S}_{cs}$}& \textbf{0.7096}& \textbf{0.8729}& \textbf{0.7863}& 0.6499& 0.8225& 0.7104 \\ 
	\end{tabular}
\end{table}

\begin{table}[t]
	\caption{Statistical evaluation results for each metric and specificity level for the comparison between our method with and without user interaction. The specific statistical test metrics for each case is denoted between square brackets.}
	\label{tab:statistical_results_interactive}
    \centering
	\begin{tabular}{c|ccc}\toprule
		\textit{Specificity} &1&2&3 \\ \midrule
 \textit{$\mathcal{N}_{w}$ $[t(152),p]$} & $[7.1,< .001]$ & $[20.3,< .001]$ & $[20.8,< .001]$ \\
 \textit{$\mathcal{R}_{fres}$ $[t(152),p]$}& $[2.5, .013]$ & $[0.3, .780]$ & $[2.1, .039]$ \\ 
  \textit{$\mathcal{S}_{cs}$ $[t(152),p]$}& $[5.3,< .001]$ & $[8,< .001]$ & $[7.7,< .001]$ \\ 
	\end{tabular}
\end{table}

\subsection{Qualitative evaluation and discussion}
The quantitative evaluation discussed earlier in this article offers a rigorous objective analysis of the strengths of our proposed methodology. However, there are some aspects of the explanations that require a more qualitative study, thus, this section discusses them.\looseness=-1

\subsubsection{Practical view on the brevity improvement}
The explanations discussed in this work would often be transmitted from robots to humans, who would need to understand and retain the explanatory content. Since our explanations are textual, humans would either listen to or read them. From state-of-the-art studies we know that to retain information, people are comfortable with a speaking pace of 150-160 words per minute (wpm), while the pace for silent reading is 250-400 wpm~\cite{doi:10.1177/1529100615623267}. Let's consider the lowest bound of those intervals, and the average explanation lengths reported in Sec.~\ref{sc:results}. For the baseline narratives, humans would spend almost \textbf{5 minutes} listening to an average narrative of specificity 2, and \textbf{12 minutes} for one of specificity 3. Our method reduces those times to \textbf{42} and \textbf{55 seconds}, respectively. In the case of silent human reading, the baseline narratives of specificity 2 and 3 would require \textbf{3} and \textbf{7 minutes}, while our explanations only \textbf{25} and \textbf{32 seconds}. This is definitely a huge advancement towards the broader goal of creating artificial agents capable of reasoning and communicating in ways that are both technically rigorous and socially acceptable, bridging the gap between formal knowledge and human-centric expression.\looseness=-1

\subsubsection{A close look at unexpected explanations}
We visually inspected the explanations generated using our approach, reading them looking for major unexpected formats or content. Our produced explanation is often a single paragraph much shorter and more clearly stated than the original ontology-based narrative. This is consistent with the objective quantitative results. However, in a few cases the explanations contain a brief summary comparing the pair of plans and some bullet points with more direct and concise major differences between the plans. This only occurs in explanations of specificity levels 2 and 3, being a bit more usual in level 3. In those cases, the explanations are still shorter and clearer than their respective ontology-based narratives. However, we found one case in which the generated explanation is really difficult to read and unusually long: \textit{Pair 27} for specificity level 3. In this case, the original ontological narrative contains 4876 words and has a readability score of -18.42. The explanation generated with our approach contains 941 words, a bad result but still shorter, and a readability score of -147.33, which is much worse than the baseline score. Our explanation includes multiple bullet points, detailing, one by one, almost all the specific differences between the pair of plans. Note that this is clearly an outlier, in fact, the second lowest readability score of our explanations is 56.44 (\textit{Pair 136}). As a major takeaway of this qualitative visual inspection, we would recommend potential users of our methodology to inspect the explanations generated for their use cases, and tune the prompts to avoid undesirable behaviors. 

\subsubsection{Discussion on explanation generation as interaction}
\AOA{Interactive explainability relies on mechanisms such as \textit{syntactic priming}, a phenomenon where speakers tend to unconsciously repeat the grammatical structure of a recently heard or read sentence. LLMs excel at syntactic priming, but they tend to generate hallucinations, often related to the semantics (meaning) of the produced text. Note that syntactic priming in humans is supported by both declarative memories (e.g. semantic and episodic memories) and also non-declarative ones~\cite{HEYSELAAR201797}. Hence, our approach deliberately combines the strengths of symbolic ontology-based episodic memories (fidelity, structure) and LLMs (adaptability, naturalness, non-declarative memory). Both components are essential to generate proper explanations. As we are interested in non-structured environments, without the adaptive component, explanations would remain static and insufficiently responsive. In the evaluation, the conducted analysis of user-driven brevity requests validates this hybrid architecture. The results demonstrate that the system successfully adapts to interaction, significantly reducing explanation length while maintaining stable readability and high semantic similarity (averaging 0.7–0.8) to the source narrative. However, we acknowledge that a user study would provide a more comprehensive analysis of this aspect of our methodology.}

\subsubsection{Discussion on explanation generation time} 
Although we did not formally report the exact generation times, we conducted informal measurements indicating that explanations take approximately 10–20 seconds to generate, including the narrative construction. This duration depends largely on the length of the original ontology-based narrative and the specific LLM and computer employed. While this response time is acceptable for offline or post-hoc explanation scenarios, it poses challenges for real-time interactive dialogues. \AOA{Note that this is not a limitation of the methodology itself, but a reflection of our computational constraints. In practice, more efficient deployment settings such as lighter models or quantized variants, can reduce latency substantially, and our methodology is compatible with such optimizations. Hence, while achieving real-time interactive latencies remains an important future direction, it is a matter of engineering optimization rather than a conceptual scientific limitation of our method.}

\section{Conclusion}
\label{sc:conclusion}
This work demonstrated the potential of integrating ontology-based reasoning with large language models (LLMs) to advance explainable human-robot interaction. By combining the logical rigor and semantic grounding of ontologies with the natural, adaptive communication abilities of LLMs, our framework enables robots to generate explanations that are both accurate and accessible to humans. The experimental validation in an industrial collaborative setting showed that this hybrid approach significantly improves the clarity and conciseness of ontology-based narratives without sacrificing their semantic depth. \AOA{Furthermore, evaluation of the interactive capability of our method offers initial positive insights that statistically demonstrate the adaptability of our approach to user interactions.} These results underscore the value of uniting structured symbolic knowledge with data-driven language generation for building transparent, trustworthy, and socially aware robotic systems. 
Future research will focus on extending this integration toward \textbf{real-time dialogue}, \textbf{multimodal explanations}, and \textbf{experiments with users} to assess the impact of such explanations on subjective perceptions of trust and collaboration. Additionally, we aim to explore how these explanation mechanisms can be used not only for human understanding but also for robot self-assessment, enabling robots to transform explanations from merely communicative tools into mechanisms for adaptive reasoning and self-improvement in \textbf{long-horizon tasks}.



\begin{acks}
This work was partially supported by the European Union under the project ARISE (HORIZON-CL4-2023-DIGITAL-EMERGING-01-101135959), and by the Spanish National Research Council (CSIC) under grant number CSIC-BILAT23120. 
\end{acks}



\bibliographystyle{ACM-Reference-Format} 
\bibliography{sample}


\end{document}